\documentclass[runningheads]{llncs}
\usepackage{graphicx}
\usepackage{comment}
\usepackage{amsmath,amssymb} % define this before the line numbering.
\usepackage{color}

\newcommand*{\our}{OSHOT\@\xspace}

\usepackage{booktabs}
\usepackage{multirow}
\usepackage{hhline}
\usepackage{xspace}
\makeatletter
\DeclareRobustCommand\onedot{\futurelet\@let@token\@onedot}
\def\@onedot{\ifx\@let@token.\else.\null\fi\xspace}
\def\eg{\emph{e.g}\onedot} 
\def\ie{\emph{i.e}\onedot}

\makeatother

\usepackage{epsfig}
\usepackage{graphicx}
\graphicspath{{./images/}}
\usepackage{overpic}
\usepackage{float}
\floatstyle{plaintop}
\restylefloat{table}
\usepackage[export]{adjustbox}
\usepackage{multirow}
\usepackage{subfig}
\usepackage{amsmath}
\DeclareMathOperator*{\argmin}{argmin}
\usepackage[ruled,vlined]{algorithm2e}
\usepackage[titletoc]{appendix}

\usepackage[font=small,labelfont=bf,labelsep=period]{caption}
\DeclareCaptionLabelFormat{andtable}{#1~#2  \&  \tablename~\thetable}

\begin{document}
\pagestyle{headings}
\mainmatter
\def\ECCVSubNumber{}  

\title{One-Shot Unsupervised Cross-Domain Detection} % Replace with your title

% CAMERA READY SUBMISSION
%\begin{comment}
\titlerunning{One-Shot Unsupervised Cross-Domain Detection}
% If the paper title is too long for the running head, you can set
% an abbreviated paper title here
%
\author{Antonio D'Innocente\inst{1,2} \and
Francesco Cappio Borlino\inst{3} \and
Silvia Bucci\inst{2,3} \and \\ 
Barbara Caputo\inst{2,3} \and
Tatiana Tommasi\inst{2,3}
}
\authorrunning{A. D'Innocente et al.}
% First names are abbreviated in the running head.
% If there are more than two authors, 'et al.' is used.
%
\institute{Sapienza University of Rome, Rome, Italy \and
Italian Institute of Technology, Turin, Italy \and
Politecnico di Torino, Turin, Italy}
%\end{comment}
%******************
\maketitle
\vspace{-4mm}
\begin{abstract}
Despite impressive progress in object detection over the last years, it is still an open challenge to reliably detect objects across visual domains. Although the topic has attracted attention recently, current approaches all rely on the ability to access a sizable amount of target data for use at training time. This is a heavy assumption, as often it is not possible to anticipate the domain where a detector will be used, nor to access it in advance for data acquisition. Consider for instance the task of monitoring  image feeds from social media: as every image is created and uploaded by a different user it belongs to a different target domain that is impossible to foresee during training. 
This paper addresses this setting, presenting an object detection algorithm able to perform unsupervised adaption across domains by using only one target sample, seen at test time. We achieve this by introducing a multi-task architecture that one-shot adapts to any incoming sample by iteratively solving a self-supervised task on it. We further enhance this auxiliary adaptation with cross-task pseudo-labeling. 
A thorough benchmark analysis against the most recent cross-domain detection methods and a detailed ablation study show the advantage of our method, which sets the state-of-the-art in the defined one-shot scenario. 

\end{abstract}

\section{Introduction}
Social media feed us every day with an unprecedented amount of visual data. Conservative estimates indicate that roughly
$10^1-10^2M$ unique images are shared everyday on Twitter,  Facebook and  Instagram. Images are uploaded by various actors, from corporations to political parties, institutions, entrepreneurs and private citizens. For the sake of freedom of expression, control over their content is limited, and their vast majority is uploaded without any textual description of their content. Their sheer magnitude makes it imperative to use algorithms to monitor, catalog and in general make sense of them, finding the right balance between protecting the privacy of citizens and their right of expression, and monitoring the spreading  of fake news (often associated with malicious intentions) while fighting illegal and hate content. This in most cases boils down to the ability to automatically associate as many tags as possible to images, which in turns means determining which objects are present in a scene.  

Object detection has been largely investigated since the infancy of computer vision  \cite{viola2001rapid,dalal2005histograms}. With the shift from shallow to deep learning, several successful algorithms have been proposed \cite{girshick2014rich,dai2016r,zhang2018single,liu2018receptive}. They mostly assume that training and test data come from the same visual domain \cite{girshick2014rich,girshick2015fast,ren2015faster}. While this is a reasonable assumption in several applications \cite{liu2016ssd,ren2015faster}, some authors have started to investigate the more challenging yet realistic scenario where the detector is trained on data from a visual source domain, and deployed at test time in a different target domain \cite{Long:2015,LongZ0J17,dcoral,Hoffman:Adda:CVPR17}. This setting, usually referred to as \emph{cross-domain detection}, heavily relies on concepts and results from the domain adaptation literature \cite{Long:2015,ganin2014unsupervised,Goodfellow:GAN:NIPS2014}. In particular, most works in cross-domain detection cast the problem in the \emph{unsupervised domain adaptation} framework \cite{inoue2018cross,Chen_2018_CVPR}: the detector has access at training time to annotated source data and unsupervised target data, from which it learns how to adapt across the two domains.  

\begin{figure}[tb]
    \centering
    \includegraphics[width=\textwidth]{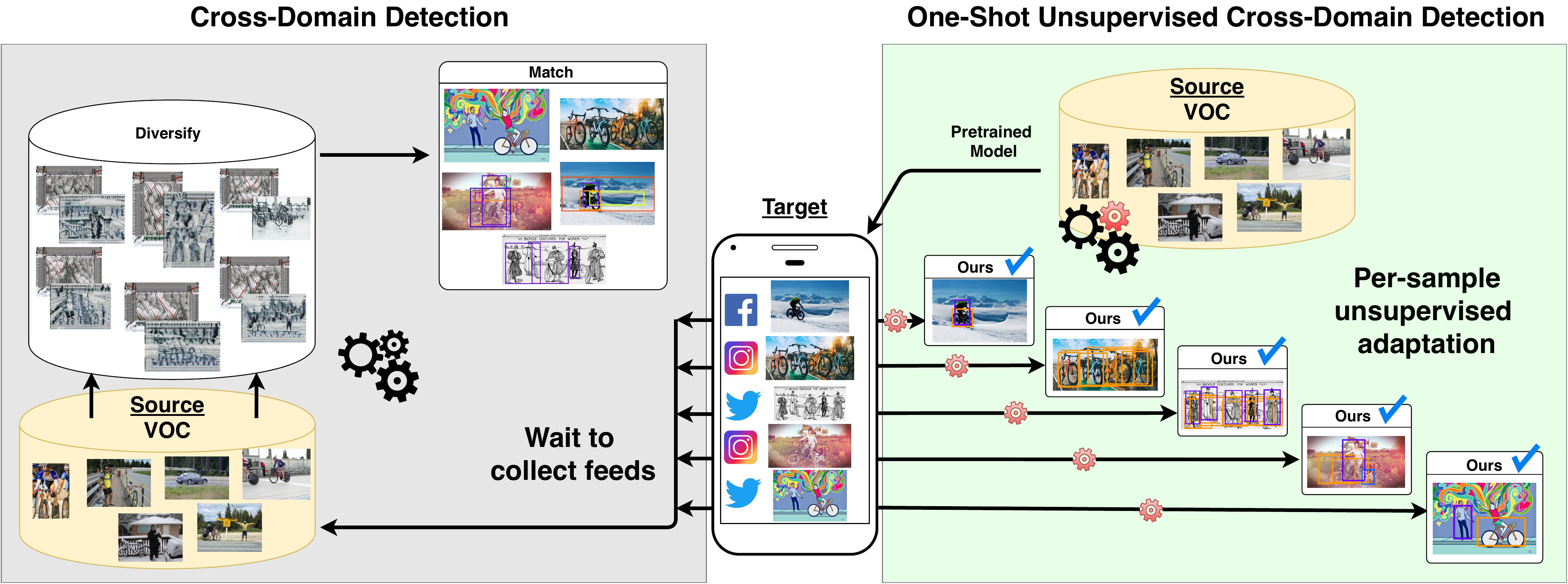}
    \caption{Each social media image comes from a different domain. Existing Cross-Domain Detection algorithms (\eg \cite{diversify&match_Kim_2019_CVPR} in the left gray box) struggle to adapt in this setting. OSHOT (right) is able to adapt across domains from one single target image, thanks to the combined use of self-supervision and pseudo-labeling
    } 
    \label{fig:scheme}\vspace{-5mm}
\end{figure}

This approach is not suitable, neither effective, for monitoring social media feeds. Consider for instance the scenario depicted in Fig \ref{fig:scheme}, where there is an incoming stream of images from various social media and the detector is asked to look for instances of the class bicycle. The images come continuously, but they are produced by different users that share them on different social platforms. Hence, even though they might contain the same object, each of them has been acquired by a different person, in a different context, under different viewpoints and illuminations --in other words, \emph{each image comes from a different visual domain, different from the visual domain where the detector has been trained}. This poses two key challenges to current cross-domain detectors: (1) to adapt to the target data, these algorithms need first to collect feeds, and only after  enough target data has been collected they can learn to adapt and start performing on the incoming images; (2) even if the algorithms have learned to adapt on target images collected from the feed up to time $t$, there is no guarantee that the images that will arrive from time $t+1$ will come from the same target domain.

This is the scenario we address.
We focus on cross-domain detection when only one target sample is available for adaptation, without any form of supervision.
We propose an object detection framework able to adapt from one target image, hence suitable for the social media scenario described above. Specifically, we build a multi task deep architecture that adapts across domains by leveraging over a pretext task.
This auxiliary knowledge is further guided by a cross-task pseudo-labeling that injects the locality specific of object detection into self-supervised learning. The result is an  architecture able to perform unsupervised adaptive object detection from a single image.  We call our method OSHOT - one shot adaptive cross-domain detection. Experiments on three different publicly available benchmarks plus a new concept database of images collected from social media clearly show the power of our method compared to previous state-of-the-art approaches.

\vspace{-5mm}\subsubsection{Contributions} To summarize, the contributions of our paper are as follows:
\begin{enumerate}
    \item we introduce the One-Shot Unsupervised Cross-Domain Detection setting, a cross-domain detection scenario where the target domain changes from sample to sample, hence adaptation can be learned only from one image. This scenario is especially relevant for monitoring social media image feeds. We are not aware of previous works addressing it.
    \item We propose OSHOT, the first cross-domain object detector able to perform one-shot unsupervised adaptation.  Our approach leverages over self-supervised one-shot learning guided by a cross-task pseudo-labeling procedure, embedded into a multi-task architecture. A thorough ablation study showcases the importance of each component.
    \item We present a new experimental setup for studying one-shot unsupervised cross-domain adaptation, designed on three existing databases plus a new test set collected from social media feed. We compare against recent algorithms in cross-domain adaptive detection \cite{Saito_2019_CVPR,diversify&match_Kim_2019_CVPR} and one-shot unsupervised learning \cite{Cohen_2019_ICCV}, achieving the state-of-the-art. 
\end{enumerate}

{}
%\clearpage

\section{Related Work}
\label{sec:related}

\subsubsection{Object Detection} Many successful object detection approaches have been developed during the past several years, starting from the original sliding window methods based on handcrafted features, till the most recent deep-learning empowered solutions.
Modern detectors can be divided into \emph{one-stage} and \emph{two-stage} techniques. In the former, classification and bounding box prediction is performed on the convolution feature map either solving a regression problem on grid cells \cite{redmon2016you} or exploiting anchor boxes at different scales and aspect ratios \cite{liu2016ssd}. In the latter, an initial stage deals with the region proposal process and is followed by a refinement stage that adjusts the coarse region localization and classify the box content.
Existing variants of this strategy differ mainly in the region proposal algorithm \cite{girshick2014rich,girshick2015fast,ren2015faster}.  
Regardless of the specific implementation, the detector robustness across visual domain remains a major issue.

\vspace{-5mm}\subsubsection{Cross-Domain Detection}
When training and test data are drawn from two different distributions a model learned on the first is doomed to fail on the second. Unsupervised domain adaptation methods attempt to close the domain gap between the annotated source on which learning is performed, and the target samples on which the model is deployed.  
Most of the literature has focused on object classification with solutions based on feature alignment  \cite{Long:2015,LongZ0J17,dcoral,hdivergence} or adversarial approaches \cite{Ganin:DANN:JMLR16,Hoffman:Adda:CVPR17}. GAN-based methods allow to directly update the visual style of the annotated source data and reduce the domain shift directly at pixel level \cite{russo17sbadagan,cycada}. 
The task of cross-domain object detection has received relatively less attention. Only in the last two years adaptive detection methods have been developed considering three main components: (i) including multiple and increasingly more accurate feature alignment modules at different internal stages, (ii) adding a preliminary pixel-level adaptation and (iii) pseudo-labeling. The last one is also known as \emph{self-training} and consists in using the output of the source model detector as coarse annotation on the target. 
The importance of considering both global and local domain adaptation,
together with a consistency regularizer to bridge the two, was first highlighted in  \cite{Chen_2018_CVPR}. The Strong-Weak (SW) method of \cite{Saito_2019_CVPR} improves over the previous one pointing out the need of a better balanced alignment with strong global and weak local adaptation and is further extended by \cite{Xie_2019_ICCV_Workshops} where the adaptive steps are multiplied at different depth in the network.
By generating new source images that look like those of the target, the Domain-Transfer (DT, \cite{inoue2018cross}) method was the first to adopt pixel adaptation for object detection and combine it with pseudo-labeling. More recently the Div-Match approach \cite{diversify&match_Kim_2019_CVPR} re-elaborated the idea of domain randomization \cite{Tobin2017DomainRF}: multiple CycleGAN \cite{CycleGAN2017} applications with different constraints produce three extra source variants with which the target can be aligned at different extent through an adversarial multi-domain discriminator.
A weak self-training procedure (WST) to reduce false negatives is combined with adversarial background score regularization (BSR) in \cite{kim2019selftraining}. 
Finally, \cite{robust_Khodabandeh_2019_ICCV} followed the pseudo-labeling strategy including an approach to deal with noisy annotations.

\vspace{-5mm}\subsubsection{Adaptive Learning on a Budget}
There is a wide literature on learning from a limited amount of data, both for classification and detection. However, in case of domain shift, learning on a target budget becomes extremely challenging. Indeed, the standard assumption for adaptive learning is that a large amount of unsupervised target samples are available at training time so that a model can capture the domain style from them and close the gap with respect to the source.
Only few attempts have been done to reduce the target cardinality. In \cite{fewshotNIPS17} the considered setting is that of \emph{few-shot supervised domain adaptation}: only a few target samples are available but they are fully labeled. In \cite{oneshotNIPS2018,Cohen_2019_ICCV} the focus is on \emph{one-shot unsupervised style transfer} with a large source dataset and a single unsupervised target image. These works propose time-costly autoencoder-based methods to generate a version of the target image that maintains its content but visually resembles the source in its global appearance. Thus the goal is image generation with no discriminative purpose. 
A related setting is that of \emph{online domain adaptation} where unsupervised target samples are initially scarce but accumulate in time \cite{Hoffman_CVPR2014,Wulfmeier2017IncrementalAD,mancini2018kitting}. In this case target samples belong to a continuous data stream with smooth domain changing, so the coherence among subsequent samples can be exploited for adaptation. 

\vspace{-5mm}\subsubsection{Self-Supervised Learning} Despite not-being manually annotated, unsupervised data is rich of structural information that can be learned by self-supervision, hiding a subpart of the data information and then trying to recover it. This procedure is generally indicated as \emph{pretext} task and possible examples are  image completion \cite{pathakCVPR16context}, colorization \cite{zhang2016colorful,larsson2017colorization}, 
relative position of patches \cite{doersch2015unsupervised,noroozi2016unsupervised}, rotation recognition \cite{gidaris2018unsupervised} and many more. Self-supervised learning has been extensively used as an initialization step for scarcely annotated supervised learning settings and very recently \cite{asano20a-critical} has shown with a thorough analysis the potential of self-supervised learning
from a single image. Several methods have also shown how self-supervision supports adaptation 
and generalization when combined with supervised learning in a multi-task framework  \cite{jigen,Bucci2019TacklingPD,Xu2019SelfsupervisedDA}.

\vspace{-5mm}\subsubsection{Our approach} for cross-domain detection relates to the described scenario of learning on a budget and exploits self-supervised learning to perform one-shot unsupervised adaptation. Specifically with OSHOT we show how to recognize objects and their location on a single target image starting from a pre-trained source model, thus without the need of accessing the source data during testing.
%\clearpage

\section{Method}
\label{sec:method}

\subsubsection{Problem Setting}
We introduce the \emph{one-shot unsupervised cross-domain detection scenario} where our goal is to predict on a single target image $x^t$, with $t$ being any target domain not available at training time, starting from $N$ annotated samples of the source domain $S=\{x^s_{i},y^s_{i}\}_{i=1}^N$. Here the structured labels $y^s=(c,b)$ describe class identity $c$ and bounding box location $b$ in each image $x^s$, and we aim to obtain $y^t$ that precisely detects objects in $x^t$ despite the domain shift.

\vspace{-5mm}\subsubsection{\our strategy} To pursue the described goal, our strategy is to train the parameters of a detection learning model such that it can be ready to get the maximal performance on a single unsupervised sample from a new domain after few gradient update steps on it. Since we have no ground truth on the target sample, we implement this strategy by learning a representation that exploits inherent data information as that captured by a \emph{self-supervised} task, and then finetune it on the target sample. 
Thus, we design our \our to include (1) an initial \emph{pretraining} phase where we extend a standard deep detection model adding an image rotation classifier, and (2) a following \emph{adaptation} stage where the network features are updated on the single target sample by further optimization of the rotation objective. 
Moreover, we exploit \emph{pseudo-labeling} to focus the auxiliary task on the local object context. A clear advantage of this solution is that we decouple source training from target testing, with no need to access the source data while adapting on the target sample.

\begin{figure*}[tb]
    \centering
\includegraphics[width=0.9\textwidth]{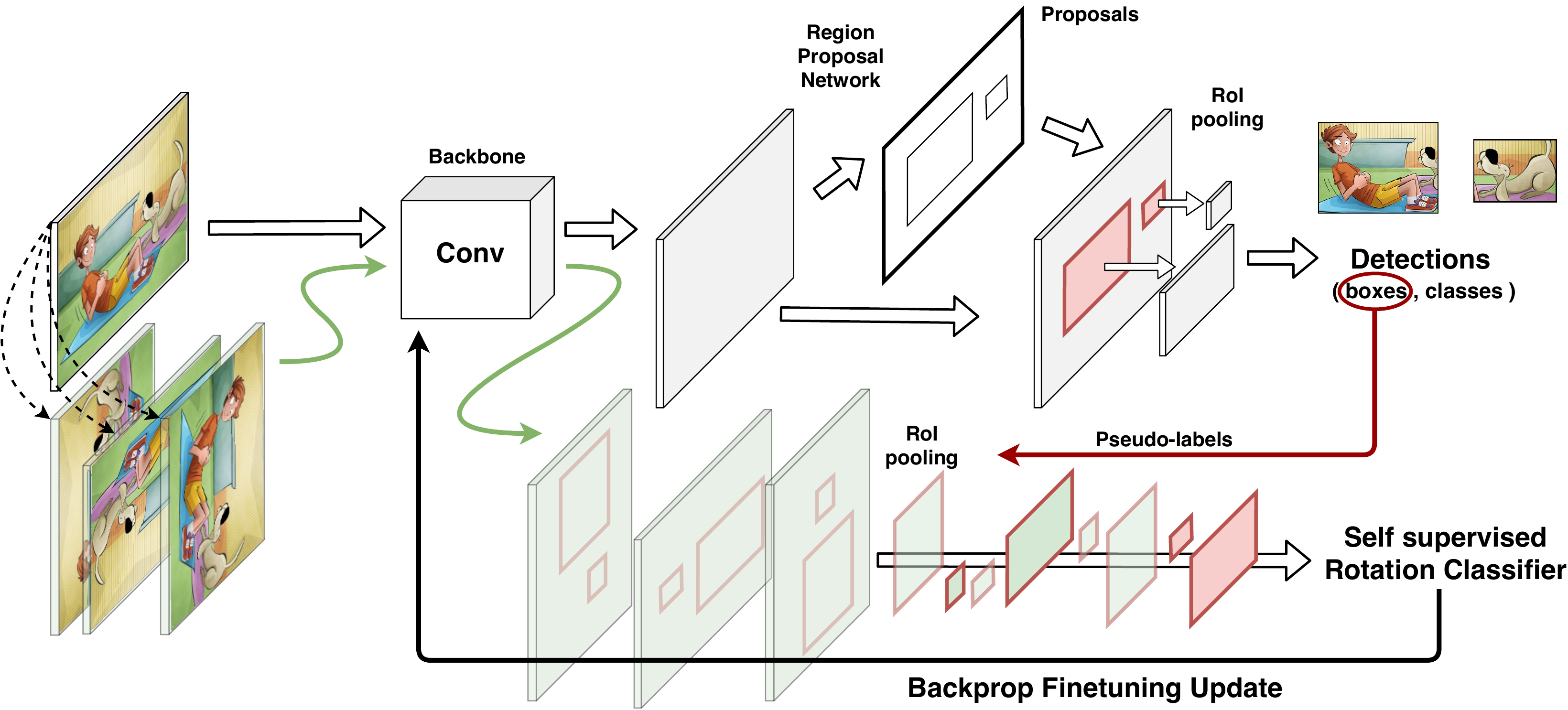}   
\caption{Visualization of the adaptive phase of \our with cross-task pseudo-labeling. 
The target image passes through the network and produces detections. While the class information is not used, the identified boxes are exploited to select object regions from the feature maps of the rotated image. The obtained region-specific feature vectors are finally sent to the rotation classifier. A number of subsequent finetuning iterations allow to adapt the convolutional backbone to the domain represented by the test image
   }
    \label{fig:fasterRCNN}\vspace{-3mm}
\end{figure*}

\vspace{-5mm}\subsubsection{Preliminaries} We leverage on Faster R-CNN \cite{ren2015faster} as our base detection model. It is a two-stage detector with three main components: an initial block of convolutional layers, a region proposal network (RPN) and a region-of-interest (ROI) based classifier.  The bottom layers transform any input image $x$ into its convolutional feature map $G_{f}(x|\theta_{f})$ where $\theta_{f}$ is used to parametrize the feature extraction model. The feature map is then used by RPN to generate candidate object proposals. Finally the ROI-wise classifier predicts the category label from the feature vector obtained using ROI-pooling. 
The training objective combines the loss of both RPN and ROI, each of them composed by two terms:
\begin{equation}
\begin{aligned}
\mathcal{L}_{d}(G_{d}(G_{f}(x|\theta_{f})|\theta_{d}), y)= & \big(\mathcal{L}_{class}(c^*) + \mathcal{L}_{regr}(b) \big)_{RPN} + \\
& \big( \mathcal{L}_{class}(c) + \mathcal{L}_{regr}(b) \big)_{ROI}~.
\end{aligned}
\end{equation}
Here $\mathcal{L}_{class}$ is a classification loss to evaluate the object recognition accuracy, while $\mathcal{L}_{regr}$ is a regression loss on the box coordinates for better localization. 
To maintain a simple notation we summarize the role of ROI and RPN with the function $G_{d}(G_{f}(x|\theta_{f})|\theta_{d})$ parametrized by $\theta_{d}$. Moreover, we use $c^*$ to highlight that RPN deals with a binary classification task to separate foreground and background objects, while ROI deals with the multi-class objective  needed to discriminate among $c$ foreground object categories. As mentioned above, ROI and RPN are applied in sequence: they both elaborate on the feature maps produced by the convolutional block, and then influence each other in the final optimization of the multi-task (classification, regression) objective function.
\vspace{-5mm}\subsubsection{\our pretraining} As a first step, we extend Faster R-CNN to include image rotation recognition. Formally, to each source training image $x^s$ we apply four geometric transformations $R(x,\alpha)$ where $\alpha= q\times90^{\circ}$ indicates rotations with $q \in \{1,\ldots,4\}$.
In this way we obtain a new set of samples $\{R(x)_j, q_j\}_{j=1}^M$ where we dropped the $\alpha$ without loss of generality.
We indicate the auxiliary rotation classifier and its parameters respectively with $G_{r}$ and $\theta_{r}$
and we train our network to optimize the following multi-task objective\\
\begin{equation}
\begin{aligned}
    \argmin_{\theta_{f}, \theta_{d}, \theta_{r}} \ & \sum_{i=1}^N\mathcal{L}_{d}(G_{d}(G_{f}(x^s_i|\theta_{f})|\theta_{d}),y^s_i) +  \lambda \sum_{j=1}^M\mathcal{L}_{r}(G_{r}(G_{f}(R(x^s)_j|\theta_{f})|\theta_{r}), q^s_j)~,
\end{aligned}
\end{equation}
where $\mathcal{L}_{r}$ is the cross-entropy loss.
When solving this problem, we can design $G_{r}$ in two different ways. Indeed it can either be a Fully Connected layer that na\"{\i}vely takes as input the feature map produced by the whole (rotated) image $G_{r}(\cdot |\theta_r) = \mbox{FC}_{\theta_r}(\cdot)$, or it can exploit the ground truth location of each object with a subselection of the features only from its bounding box in the original map  
$G_{r}(\cdot |\theta_r) = \mbox{FC}_{\theta_r}(boxcrop(\cdot))$. The $boxcrop$ operation includes pooling to rescale the feature dimension before entering the final FC layer. 
In this last case the network is encouraged to focus only on the object orientation without introducing noisy information from the background and provides better results with respect to the whole image option as we discuss in section \ref{sec:ablation}.
In practical terms, both in the case of image and box rotations, we randomly pick one rotation angle per instance, rather than considering all four of them: this avoids any troublesome unbalance between rotated and non-rotated data when solving the multi-task optimization problem.

\vspace{-5mm}\subsubsection{\our adaptation} Given the single target image $x^t$, we fine-tune the backbone's parameters $\theta_{f}$ by iteratively solving a self-supervised task on it. 
This allows to adapt the original feature representation both to the content and to the style of the new sample. Specifically, we start from the rotated versions $R(x^t)$  of the provided sample and optimize the rotation classifier through 
\begin{equation}
    \argmin_{\theta_{f}, \theta_{r}}  \mathcal{L}_{r}(G_{r}(G_{f}(R(x^t)|\theta_{f})|\theta_{r}),q^t)~.
    \label{eq:finetuning}
\end{equation}
This process involves only $G_{f}$ and $G_{r}$, while the RPN and ROI detection components described by $G_{d}$ remain unchanged. In the following we use $\gamma$ to indicate the number of gradient steps (\ie iterations), with $\gamma=0$ corresponding to the \our{} pretraining phase. At the end of the finetuning process, the inner feature model is  described by $\theta^*_f$ and the detection prediction on $x^t$ is obtained by $y^{t*} = G_{d}(G_{f}(x^t|\theta^*_{f})|\theta_{d})$.

\vspace{-5mm}\subsubsection{Cross-task pseudo-labeling}
As in the pretraining phase, also at this stage we have two possible choices to design $G_{r}$: either considering the whole feature map $G_{r}(\cdot |\theta_r) = \mbox{FC}_{\theta_r}(\cdot)$, or focusing on the object locations $G_{r}(\cdot |\theta_r) = \mbox{FC}_{\theta_r}(pseudoboxcrop(\cdot))$. For both variants we include dropout to prevent overfitting on the single target sample. With $pseudoboxcrop$ we mean a localized feature extraction operation analogous to that discussed in the previous paragraph, but obtained through a particular form of \emph{cross-task self-training}. 
Specifically we follow the self-training strategy used in  \cite{kim2019selftraining,inoue2018cross} with a cross-task variant: instead of reusing the pseudo-labels produced by the source model on the target to update the detector, we exploit them for the self-supervised rotation classifier.
In this way we keep the advantage of the self-training initialization, largely reducing the risks of error propagation due to wrong pseudo-labels.

More practically, we start from the $(\theta_{f},\theta_{d})$ model parameters of the pretraining stage and we get the feature maps from all the rotated version of the target sample $[G_{f}(\{R(x^t),q\}|\theta_{f}),q={1,\ldots,4}]$. 
Only the feature map produced by the original image (\ie $q=4$) is provided as input to the RPN and ROI network components to get the predicted detection $y^{t}=(c,b)=G_{d}(G_{f}(x^t|\theta_{f})|\theta_{d})$. This pseudo-label is composed by the class label $c$ and the bounding box location $b$. We discard the first and consider only the second to localize the region containing an object in all the four feature maps, also recalibrating the position to compensate for the orientation of each map. Once passed through this \emph{pseudoboxcrop} operation the obtained features are used to finetune the rotation classifier, updating the bottom convolutional network block.

\section{Experiments}

\subsection{Datasets}
\label{sec:datasets}

\begin{figure}[tb]
    \centering
    \includegraphics[width=\textwidth]{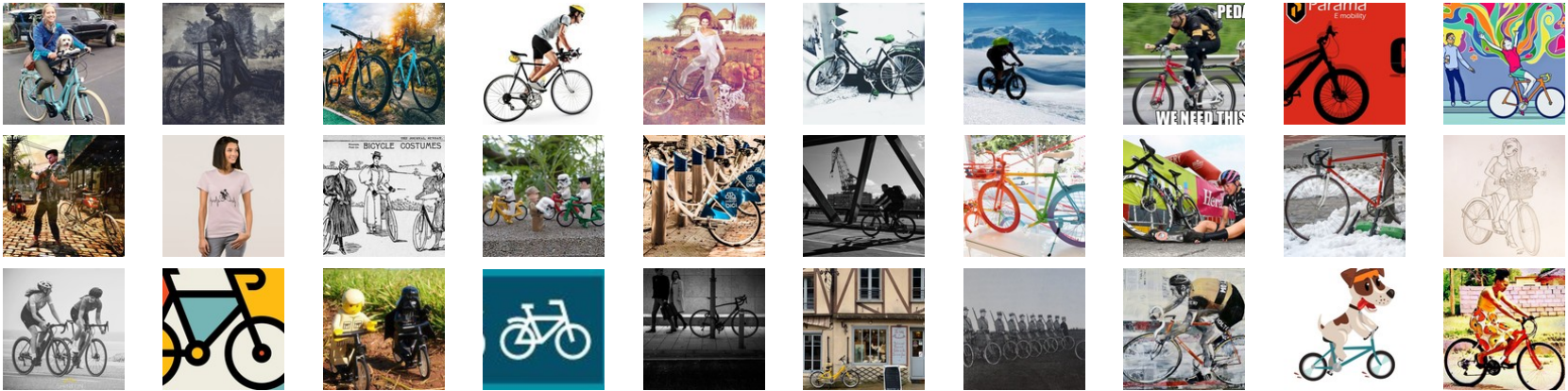}
    \caption{The Social Bikes concept-dataset. We can see how a random data acquisition from multiple users/feeds leads to a target distribution with several, uneven domain shifts}
    \label{fig:socialbikes}\vspace{-3mm}
\end{figure}{}
\subsubsection{Real-World (VOC)} Pascal-VOC \cite{everingham2010pascal} is the standard real-world image dataset for object detection benchmarks. VOC2007 and VOC2012 both contain bounding boxes annotations of 20 common categories. VOC2007 has 5011 images in the train-val split and 4952 images in the test split, while VOC2012 contains 11540 images in the train-val split.

\vspace{-5mm}\subsubsection{Artistic Media Datasets (AMD)} Clipart1k, Comic2k and Watercolor2k \cite{inoue2018cross} are three object detection datasets designed for benchmarking Domain Adaptation methods when the source domain is Pascal-VOC. Clipart1k shares its 20 categories with Pascal-VOC, and has 500 images in the training set and 500 images in the test set. Comic2k and Watercolor2k both have the same 6 classes (a subset of the 20 classes of Pascal-VOC), and 1000-1000 images in the training-test splits each.
\vspace{-5mm}\subsubsection{Cityscapes} Cityscapes \cite{cordts2016cityscapes} is an urban street scene dataset with pixel level annotations of 8 categories. It has 2975 images in the training split and 500 images in the validation split. Since this dataset doesn't have bounding boxes annotations, we use the instance level pixel annotations to generate bounding boxes of objects, as in \cite{Chen_2018_CVPR}.
\vspace{-5mm}\subsubsection{Foggy Cityscapes} \cite{sakaridis2018semantic} is an urban street dataset obtained by adding different levels of synthetic fog to Cityscapes images. We only consider images with the highest amount of artificial fog, thus training-validation split have 2975-500 images respectively.
\vspace{-5mm}\subsubsection{KITTI} \cite{kitti} is a dataset designed for use in mobile robotics and autonomous driving research. Following \cite{Chen_2018_CVPR}, we use the full 7481 images for both training (when used as source) and evaluation (when used as target).
\vspace{-4mm}\subsubsection{Social Bikes} is our new concept-dataset containing 30 images of scenes with persons/bicycles collected from \textit{Twitter}, \textit{Instagram} and \textit{Facebook} by searching for \emph{\#bike} tags. Square crops of the full dataset are shown in figure \ref{fig:socialbikes}: it is clear how images acquired randomly from social feeds possess diverse style properties and cannot be grouped under a single shared domain. %This concept-dataset shows how images  

\subsection{Performance analysis}
\label{sec:setup}

\subsubsection{Experimental Setup} We evaluate \our on several testbeds. We start from the VOC$\rightarrow$Social Bikes transfer as a proof of concept experiment. Moreover, we consider the standard cross-domain benchmarks 
VOC $\rightarrow$ Clipart1k, VOC $\rightarrow$ Comic2k, VOC $\rightarrow$ Watercolor2k, Cityscapes $\rightarrow$ FoggyCityscapes, KITTI $\rightarrow$ Cityscapes and Cityscapes $\rightarrow$ KITTI with the added constraint of never using the target data during training.
Our base detector is Faster-RCNN\footnote{PyTorch-based implementation \cite{massa2018mrcnn}.} with a ResNet-50 \cite{he2016deep} backbone pre-trained on ImageNet, region proposal network with 300 top proposals after non-maximum-supression, and anchors at three scales (128, 256, 512) and three aspect ratios (1:1, 1:2, 2:1).

\vspace{-3mm}\paragraph{\our pretraining} We always resize the image's shorter size to 600 pixels and apply random horizontal flipping. Unless differently specified, we train the base network for 70k iterations using SGD with momentum set at $0.9$, the initial learning rate is set at $0.001$ and decayed after 50k iterations. We use a batch size of 1, keep batch normalization layers fixed for both pretraining and adaptation phases and freeze the first 2 blocks of ResNet50. The weight of the auxiliary task is set to $\lambda=0.05$.

\vspace{-3mm}\paragraph{\our adaptation} We increase the weight of the auxiliary task to $\lambda=0.2$ to speed up adaptation and keep all other training hyperparameters fixed. For \emph{each} test instance, we finetune the \emph{initial} model on the auxiliary task for 30 iterations before testing. 

\vspace{-3mm}\paragraph{Benchmark methods} We compare \our with the following algorithms. \emph{FRCNN}: baseline Faster-RCNN with ResNet50 backbone, trained on the source domain and deployed on the target without further adaptation. \emph{DivMatch} \cite{diversify&match_Kim_2019_CVPR}:
cross-domain detection algorithm that, by exploiting target data, creates multiple randomized domains via CycleGAN and aligns their representations using an adversarial loss. \emph{SW} \cite{Saito_2019_CVPR}: adaptive detection algorithm that aligns source and target features based on global context similarity. For both DivMatch and SW, we use a ResNet-50 backbone pretrained on ImageNet for fair comparison. Since all cross-domain algorithms need target data in advance and are not designed to work in our one-shot unsupervised setting, we provide them with the advantage of 10 target images accessible during training and collect average precision statistics during inference under the favorable assumption that the target domain will not shift after deployment.

\begin{table}[t]
\centering
\begin{tabular}{cc}
\begin{adjustbox}{width=0.4\textwidth}
%\centering
\begin{tabular}{c@{~~}c@{~~}c@{~~}|c@{~~}}
\hline%\hline
\multicolumn{4}{c}{\textsl{\textbf{One-Shot} Target}}\\
\hline
Method & person & bicycle & mAP\\ \hline
FRCNN &  67.7 & 56.6 & 62.1 \\ \hline
\textbf{\textit{\our}} ($\gamma = 0$) & 72.1 & 52.8 & 62.4 \\
\textbf{\textit{\our}} ($\gamma = 30$) & 69.4 & 59.4 & \textbf{64.4}\\ \hline%\hline
\multicolumn{4}{c}{\textsl{\textbf{Full} Target}}\\
\hline
DivMatch  \cite{diversify&match_Kim_2019_CVPR} & 63.7 & 51.7 & 57.7 \\
SW \cite{Saito_2019_CVPR}  & 63.2 & 44.3 & 53.7 \\\hline 
\end{tabular}
\end{adjustbox}
&
\qquad
\centering
\includegraphics[width=0.4\textwidth,valign=m]{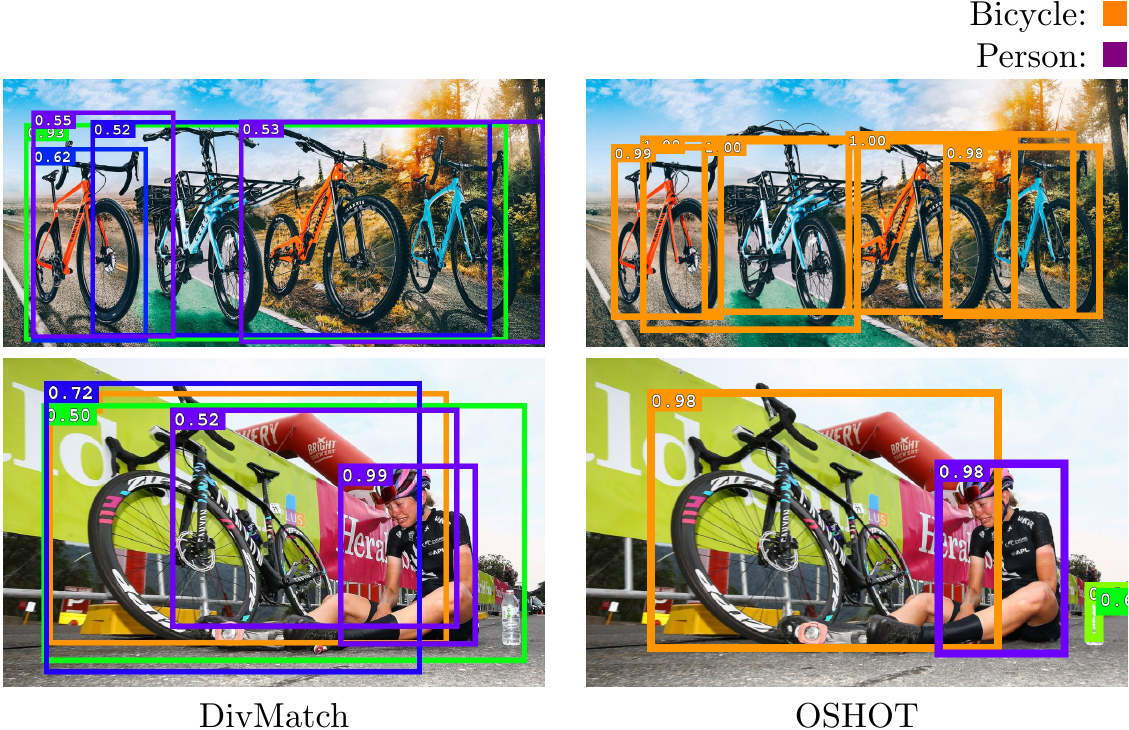}
\end{tabular}
\caption{VOC $\rightarrow$ Social Bikes results and visualization of DivMatch (left) and OSHOT (right) detections. Numbers associated with bounding boxes indicate the model's confidence in localization. Examples show how OSHOT detection is accurate, while most DivMatch boxes are false positives}
\label{table:social}\vspace{-3mm}
\end{table}

\begin{comment}
\begin{table*}[tb]
\centering
\begin{adjustbox}{width=0.4\textwidth}
\centering
\begin{tabular}{c@{~~}c@{~~}c@{~~}|c@{~~}}
\hline\hline
\multicolumn{4}{c}{\textsl{\textbf{One-Shot} Target}}\\
\hline
Method & person & bicycle & mAP\\ \hline
%\multicolumn{22}{c}{\textbf{Source Only}}\\
%\hline
FRCNN &  67.7 & 56.6 & 62.1 \\ \hline
\textbf{\textit{\our}} ($\gamma = 0$) & 72.1 & 52.8 & 62.4 \\
\textbf{\textit{\our}} ($\gamma = 30$) & 69.4 & 59.4 & \textbf{64.4}\\ \hline\hline
\multicolumn{4}{c}{\textsl{\textbf{Full} Target}}\\
\hline
DivMatch  \cite{diversify&match_Kim_2019_CVPR} & 63.7 & 51.7 & 57.7 \\
SW \cite{Saito_2019_CVPR}  & 63.2 & 44.3 & 53.7 \\\hline 
\end{tabular}
\end{adjustbox}

\caption{VOC $\rightarrow$ Social Bikes}
\label{table:social}\vspace{-3mm}
\end{table*}

\end{comment}

\vspace{-5mm}\subsubsection{Adapting to social feeds} When data is collected from multiple sources, the  assumption that all target images originate from the same underlying distribution does not hold and standard cross-domain detection methods are penalized regardless of the number of seen target samples. We pretrain the source detector on Pascal VOC, and deploy it on Social Bikes. We consider only the bicycle and person annotations for this target, since all other instances of VOC classes are scarce.
\vspace{-4mm}\paragraph{Results} We report results in table \ref{table:social}. \our outperforms all proposed counterparts, with a mAP score of $64.4$. Despite granting them the full target, adaptive algorithms incur in negative transfer due to data scarcity and large variety of target styles.

\vspace{-5mm}\subsubsection{Large distribution shifts} Artistic images are difficult benchmarks for cross-domain methods. Unpredictable perturbations in shapes and colors are challenging to detectors trained only on realistic images, for which labeled data is more readily available. Here, we investigate the effectiveness of our one-shot transfer to artistic domains by training the source detector on Pascal VOC an deploying it on Clipart, Comic and Watercolor datasets.  
\vspace{-3mm}\paragraph{Results} Table \ref{table:VOC2AMD} summarizes results on the three adaptation splits. We can see how \our with 30 finetuning iterations outperforms all competitors, gaining mAP increases ranging from $7.5$ points on Clipart to $9.2$ points on Watercolor. Cross-detection methods perform poorly in this setting, despite using 9 more samples in the adaptation phase compared to \our that only uses the test sample. These results confirm that they are not designed to tackle data scarcity conditions and exhibit negligible improvements compared to the baseline.

\begin{table*}[tb]
\centering
\begin{adjustbox}{width=1\textwidth}
\subfloat[VOC $\rightarrow$ Clipart]{{}\begin{tabular}{c@{~~}c@{~~}c@{~~}c@{~~}c@{~~}c@{~~}c@{~~}c@{~~}c@{~~}c@{~~}c@{~~}c@{~~}c@{~~}c@{~~}c@{~~}c@{~~}c@{~~}c@{~~}c@{~~}c@{~~}c@{~~}|c@{~~}}
\hline%\hline
\multicolumn{22}{c}{\textsl{\textbf{One-Shot} Target}}\\
\hline
%\hline
Method & aero & bike & bird & boat & bottle & bus & car & cat & chair & cow & table & dog & horse & mbike & person & plant & sheep & sofa & train & tv & mAP\\ \hline
%\multicolumn{22}{c}{\textbf{Source Only}}\\
%\hline
FRCNN & 18.5 & 43.3 & 20.4 & 13.3 & 21.0 & 47.8 & 29.0 & 16.9 & 28.8 & 12.5 & 19.5 & 17.1 & 23.8 & 40.6 & 34.9 & 34.7 & 9.1 & 18.3 & 40.2 & 38.0 & 26.4 \\ \hline
\textbf{\textit{\our}} ($\gamma = 0$) & 23.1 & 55.3 & 22.7 & 21.4 & 26.8 & 53.3 & 28.9 & 4.6 & 31.4 & 9.2 & 27.8 & 9.6 & 30.9 & 47.0 & 38.2 & 35.2 & 11.1 & 20.4 & 36.0 & 33.6 & 28.3 \\
\textbf{\textit{\our}} ($\gamma = 10$) & 25.4 & 61.6 & 23.8 & 21.1 & 31.3 & 55.1 & 31.6 & 5.3 & 34.0 & 10.1 & 28.8 & 7.3 & 33.1 & 59.9 & 44.2 & 38.8 & 15.9 & 19.1 & 39.5 & 33.9 & 31.0\\
\textbf{\textit{\our}} ($\gamma = 30$) & 25.4 & 56.0 & 24.7 & 25.3 & 36.7 & 58.0 & 34.4 & 5.9 & 34.9 & 10.3 & 29.2 & 11.8 & 46.9 & 70.9 & 52.9 & 41.5 & 21.1 & 21.0 & 38.5 & 31.8 & \textbf{33.9}\\ \hline\hline
\multicolumn{22}{c}{\textsl{\textbf{Ten-Shot} Target}}\\
\hline
DivMatch  \cite{diversify&match_Kim_2019_CVPR} & 19.5 & 57.2 & 17.0 & 23.8 & 14.4 & 25.4 & 29.4 & 2.7 & 35.0 & 8.4 & 22.9 & 14.2 & 30.0 & 55.6 & 50.8 & 30.2 & 1.9 & 12.3 & 37.8 & 37.2 & 26.3 \\ 
SW \cite{Saito_2019_CVPR} & 21.5 & 39.9 & 21.7 & 20.5 & 32.7 & 34.1 & 25.1 & 8.5 & 33.2 & 10.9 & 15.2 & 3.4 & 32.2 & 56.9 & 46.5 & 35.4 & 14.7 & 15.2 & 29.2 & 32.0 & 26.4 \\ \hline
\end{tabular}}
\end{adjustbox}
\begin{adjustbox}{width=1\textwidth}
\subfloat[VOC $\rightarrow$ Comic]{\begin{tabular}{c@{~~}c@{~~}c@{~~}c@{~~}c@{~~}c@{~~}c@{~~}|c@{~~}}
\hline%\hline
\multicolumn{8}{c}{\textsl{\textbf{One-Shot} Target}}\\
\hline
Method & bike & bird & car & cat & dog & person &  mAP\\ \hline
FRCNN & 25.2 & 10.0 & 21.1 & 14.1 & 11.0 & 27.1 & 18.1 \\ \hline
\textbf{\textit{\our}} ($\gamma = 0$) & 26.9 & 11.6 & 22.7 & 9.1 & 14.2 & 28.3 & 18.8 \\
\textbf{\textit{\our}} ($\gamma = 10$) & 35.5 & 11.7 & 25.1 & 9.1 & 15.8 & 34.5 & 22.0 \\
\textbf{\textit{\our}} ($\gamma = 30$) & 35.2 & 14.4 & 30.0 & 14.8 & 20.0 & 46.7 & \textbf{26.9} \\  \hline\hline
\multicolumn{8}{c}{\textsl{\textbf{Ten-Shot} Target}}\\
\hline
DivMatch  \cite{diversify&match_Kim_2019_CVPR} & 27.1 & 12.3 & 26.2 & 11.5 & 13.8 & 34.0 & 20.8 \\ 
SW \cite{Saito_2019_CVPR} & 21.2 & 14.8 & 18.7 & 12.4 & 14.9 & 43.9 & 21.0 \\ \hline
\end{tabular}} \hspace{0.5cm}
\subfloat[VOC $\rightarrow$ Watercolor]{\begin{tabular}{c@{~~}c@{~~}c@{~~}c@{~~}c@{~~}c@{~~}c@{~~}|c@{~~}}
\hline%\hline
\multicolumn{8}{c}{\textsl{\textbf{One-Shot} Target}}\\
\hline
Method & bike & bird & car & cat & dog & person &  mAP\\ \hline
FRCNN & 62.5 & 39.7 & 43.4 & 31.9 & 26.7 & 52.4 & 42.8 \\ \hline
\textbf{\textit{\our}} ($\gamma = 0$) & 70.2 & 46.7 & 45.5 & 31.2 & 27.2 & 55.7 & 46.1\\
\textbf{\textit{\our}} ($\gamma = 10$) & 70.2 & 46.7 & 48.1 & 30.9 & 32.3 & 59.9 & 48.0\\
\textbf{\textit{\our}} ($\gamma = 30$) & 77.1 & 44.7 & 52.4 & 37.3 & 37.0 & 63.3 & \textbf{52.0}\\ \hline\hline
\multicolumn{8}{c}{\textsl{\textbf{Ten-Shot} Target}}\\
\hline
DivMatch  \cite{diversify&match_Kim_2019_CVPR} & 64.6 & 44.1 & 44.6 & 34.1 & 24.9 & 60.0 & 45.4 \\
SW \cite{Saito_2019_CVPR}  & 66.3 & 41.1 & 41.1 & 30.5 & 20.5 & 52.3 & 42.0 \\
\hline
\end{tabular}}
\end{adjustbox}
\vspace{-4mm}
\caption{VOC $\rightarrow$ AMD}
\label{table:VOC2AMD}
\end{table*}

\vspace{-5mm}\subsubsection{Adverse weather} Low level domain shifts occur when weather changes from training to testing. Some peculiar environmental conditions, such as fog, may be disregarded in source data acquisition,
yet adaptation to these circumstances is crucial for real world applications.
We assess the performance of \our on Cityscapes $\rightarrow$ FoggyCityscapes. We train our base detector on Cityscapes for 30k iterations without stepdown, as in \cite{cai2019exploring}. We select the best performing model on the Cityscapes validation split and deploy it to FoggyCityscapes. 
\vspace{-4mm}\paragraph{Results} Experimental evaluation in table \ref{table:C2F} shows that \our outperforms all compared approaches. Without finetuning iterations, performance using the auxiliary rotation task increases compared to the baseline. Subsequent finetuning iterations on the target sample improve these results, and 30 iterations yield models able to outperform the second-best method by $5$ mAP. Cross-domain algorithms used in this setting struggle to surpass the baseline (DivMatch) or suffer negative transfer (SW).

\begin{table*}[]
\centering
\begin{adjustbox}{width=0.7\textwidth}
\centering
\begin{tabular}{c@{~~}c@{~~}c@{~~}c@{~~}c@{~~}c@{~~}c@{~~}c@{~~}c@{~~}|c@{~~}}
\hline
\multicolumn{10}{c}{\textsl{\textbf{One-Shot} Target}}\\
\hline
Method & person & rider & car & truck & bus & train & mcycle & bicycle & mAP\\ \hline
%\multicolumn{22}{c}{\textbf{Source Only}}\\
%\hline
FRCNN & 30.4 & 36.3 & 41.4 & 18.5  & 32.8 & 9.1 & 20.3 & 25.9 & 26.8 \\ \hline
\textbf{\textit{\our}} ($\gamma = 0$) & 31.8 & 42.0 & 42.6 & 20.1 & 31.6 & 10.6 & 24.8 & 30.7 & 29.3 \\
\textbf{\textit{\our}} ($\gamma = 10$) & 31.9 & 41.9 & 43.0 & 19.7 & 38.0 & 10.4 & 25.5 & 30.2 & 30.1\\
\textbf{\textit{\our}} ($\gamma = 30$) & 32.1 & 46.1 & 43.1 & 20.4 & 39.8 & 15.9 & 27.1 & 32.4 & \textbf{31.9}\\ \hline\hline
\multicolumn{10}{c}{\textsl{\textbf{Ten-Shot} Target}}\\
\hline
DivMatch  \cite{diversify&match_Kim_2019_CVPR} & 27.6 & 38.1 & 42.9 & 17.1 & 27.6 & 14.3 & 14.6 & 32.8 & 26.9 \\
SW \cite{Saito_2019_CVPR}  & 25.5 & 30.8 & 40.4 & 21.1 & 26.1 & 34.5 & 6.1 & 13.4 & 24.7 \\\hline 
\end{tabular}
\end{adjustbox}
%\vspace{2mm}
\caption{Cityscapes $\rightarrow$ FoggyCityscapes}
\label{table:C2F}\vspace{-3mm}
\end{table*}

\vspace{-5mm}\subsubsection{Cross-camera transfer} Dataset bias between training and testing are unavoidable in practical applications.
Subtle changes in illumination conditions and camera resolution might preclude a model trained on one realistic domain to optimally perform in another realistic but different domain. We test adaptation between KITTI and Cityscapes in both directions. For cross-domain evaluation we consider only the label car as standard practice.
\vspace{-4mm}\paragraph{Results} In table \ref{table:KITTI}, \our improves by $7$ mAP points on KITTI $\rightarrow$ Cityscapes compared to the FRCNN baseline. DivMatch and SW both show a gain in this split, with SW obtaining the highest mAP of $39.2$ in the ten-shot setting. This is not surprising however, Cityscapes has low inter-domain variance, as shown in the visualization of table \ref{table:KITTI}, therefore cross-domain methods perform well even with few target samples if the distribution doesn't change after adaptation. In Cityscapes $\rightarrow$ KITTI, adaptation performance for all methods is similar, with \our with $\gamma=0$ obtaining the highest mAP of $75.4$. The Faster-RCNN baseline on KITTI scores an high starting mAP of $75.1$ and, in this favorable condition, detection doesn't benefit from adaptation.

\begin{table}[tb]
\centering
\begin{tabular}{cc}
\begin{adjustbox}{width=0.45\textwidth}
%\centering
\begin{tabular}{c@{~~}c@{~~}c@{~~}c@{~~}c@{~~}c@{~~}c@{~~}|c@{~~}}
\hline
\multicolumn{3}{c}{\textsl{\textbf{One-Shot} Target}}\\
\hline
Method & KITTI $\rightarrow$ Cityscapes & Cityscapes $\rightarrow$ KITTI\\ \hline
%\multicolumn{22}{c}{\textbf{Source Only}}\\
%\hline
FRCNN & 26.5 & 75.1 \\ \hline
\textbf{\textit{\our}} $\gamma = 0$ & 26.2 & \textbf{75.4} \\
\textbf{\textit{\our}} $\gamma = 10$ & 33.2 & 75.3 \\
\textbf{\textit{\our}} $\gamma = 30$ & \textbf{33.5} & 75.0 \\ \hline\hline
\multicolumn{3}{c}{\textsl{\textbf{Ten-Shot} Target}}\\
\hline
DivMatch  \cite{diversify&match_Kim_2019_CVPR} & 37.9 & 74.1 \\ 
SW \cite{Saito_2019_CVPR}  & 39.2 & 74.6 \\ \hline 
\end{tabular}
\end{adjustbox}
&
\qquad
\centering
\includegraphics[width=0.4\textwidth,valign=m]{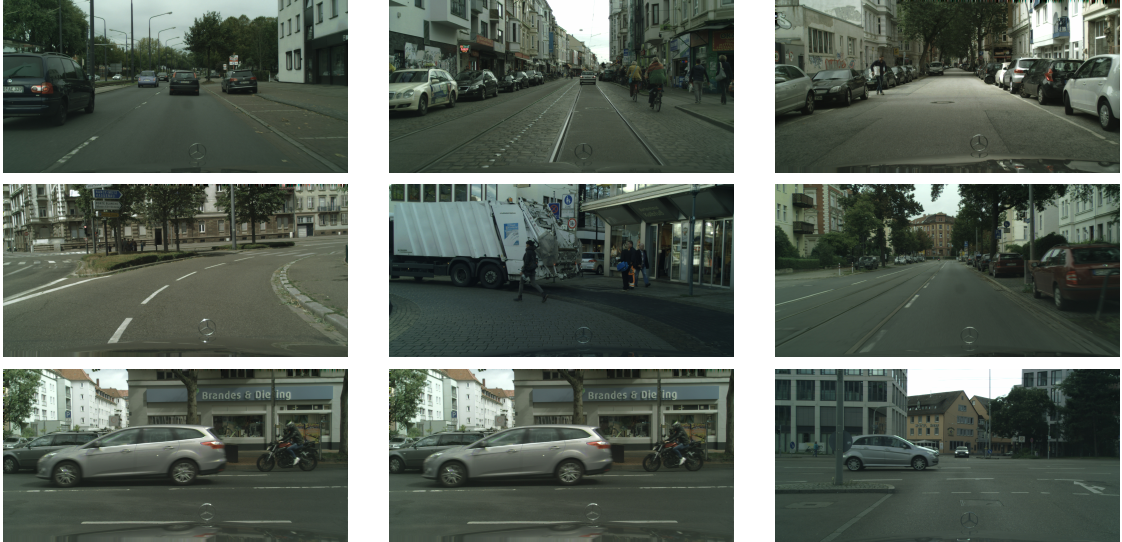}
\end{tabular}
\caption{mAP of 'car' class in KITTI/Cityscapes detection transfers}
\label{table:KITTI}
\end{table}

\begin{comment}
\begin{table*}[tb]
\centering
\begin{adjustbox}{width=0.5\textwidth}
\begin{tabular}{c@{~~}c@{~~}c@{~~}c@{~~}c@{~~}c@{~~}c@{~~}|c@{~~}}
\hline\hline
\multicolumn{3}{c}{\textsl{\textbf{One-Shot} Target}}\\
\hline
Method & KITTI $\rightarrow$ Cityscapes & Cityscapes $\rightarrow$ KITTI\\ \hline
%\multicolumn{22}{c}{\textbf{Source Only}}\\
%\hline
FRCNN & 26.5 & 75.1 \\ \hline
FRCNN-R & 26.2 & \textbf{75.4} \\
\textbf{\textit{\our}} $\lambda = 10$ & 33.2 & 75.3 \\
\textbf{\textit{\our}} $\lambda = 30$ & \textbf{33.5} & 75.0 \\ \hline\hline
\multicolumn{3}{c}{\textsl{\textbf{Ten-Shot} Target}}\\
\hline
DivMatch  \cite{diversify&match_Kim_2019_CVPR} & 37.9 & 74.1 \\ 
SW \cite{Saito_2019_CVPR}  & 39.2 & 74.6 \\ \hline 
\end{tabular}
\end{adjustbox}
%\vspace{2mm}
\caption{mAP of 'car' class in KITTI/Cityscapes detection transfers \textbf{missing some numbers}}
\label{table:KITTI}
\end{table*}\vspace{-3mm}
\end{comment}

\subsection{Comparison with One-Shot Style Transfer}
\label{sec:oneshotstyle}

Although not specifically designed for cross-domain detection, in principle it is possible to apply one-shot style transfer methods as an alternative solution for our setting. We use BiOST \cite{Cohen_2019_ICCV}, the current state-of-the-art method for one-shot transfer, to modify the style of the target sample towards that of the source domain before performing inference. Due to the time-heavy requirements to perform BiOST on each test sample \footnote{The one-shot translation of \cite{Cohen_2019_ICCV} requires the training of a double-variational autoencoder using the entire source training set plus the target sample. Through personal communication with the author, we fix the length of this training to 5 epochs.}, we test it on Social Bikes and on a random subset of 100 Clipart images that we name Clipart100. We compare performance and time requirements of \our and BiOST on these two targets. Speed has been computed on an RTX2080Ti with full precision settings.

\begin{table*}[tb]
\centering
\begin{adjustbox}{width=0.7\textwidth}
\begin{tabular}{c@{~~}c@{~~}c@{~~}c@{~~}}
\hline
 & FRCNN & BiOST \cite{Cohen_2019_ICCV} & OSHOT ($\gamma=30$)\\ \hline
%\multicolumn{22}{c}{\textbf{Source Only}}\\
%\hline
mAP on Clipart100 & 27.9 & 29.8 & \textbf{30.7} \\ \hline
mAP on Social Bikes & 62.1 & 51.1 & \textbf{64.4} \\ \hline
Adaptation time (seconds per sample) & - & $\sim 2.4*10^{4}$ & 7.8 \\
\hline
\end{tabular}
\end{adjustbox}
\caption{Comparison between baseline, one-shot syle transfer and OSHOT in the one-shot unsupervised cross-domain detection setting}
\label{table:biost}\vspace{-3mm}
\end{table*}

\vspace{-4mm}\paragraph{Results} Table \ref{table:biost} shows summary mAP results using BiOST and \our. On Clipart100, the baseline Faster-RCNN detector obtains $27.9$ mAP. We can see how BiOST is effective in the adaptation from one-sample, gaining $1.9$ points over the baseline, however it is outperformed by \our, which obtains $30.7$ mAP. On Social Bikes, while \our still outperforms the baseline, BiOST incurs in negative transfer, indicating that it was not able to effectively modify the source's style on the images we collected. Furthermore, BiOST is affected by two strong issues: (1) it has an extremely high one-shot translation time, that requires more than 6 hours to modify the style of a single source instance, and (2) it works under the strict assumption of having the entire source training set available at any time to perform the OST step. Due to these weaknesses, and the fact that \our still outperforms BiOST, we argue that existing one-shot translation methods are not suitable for one shot unsupervised cross-domain adaptation.

\subsection{Ablation Study}
\label{sec:ablation}

\subsubsection{Detection error analysis} 
\begin{figure}[]
    \centering
    \includegraphics[width=0.9\textwidth]{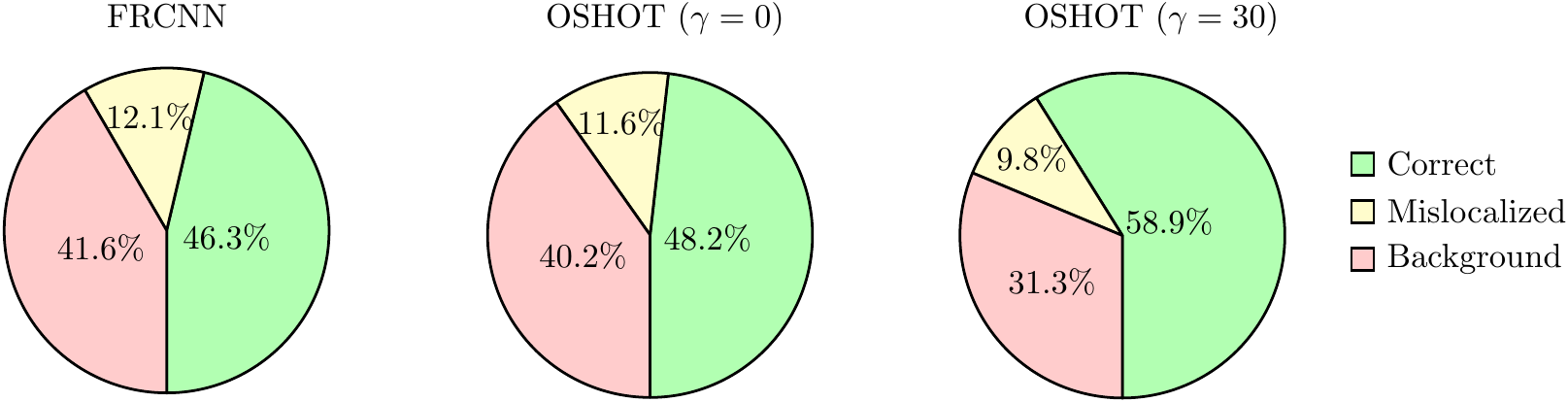}
    \caption{Detection error analysis on the most confident detections on Clipart}
    \label{fig:error}
\end{figure}

Following \cite{hoiem2012diagnosing}, we provide detection error analysis for VOC $\rightarrow$ Clipart setting in figure \ref{fig:error}. We select the 1000 most confident detections, and assign error classes based on IoU with ground truth (IoUgt). Errors are categorized in three types: correct (IoUgt $\geqslant$ 0.5), mislocalized (0.3 $\leqslant$ IoUgt $<$ 0.5) and background (IoUgt $<$ 0.3). Results show that, compared to the baseline FRCNN model, the regularization effect of adding a self-supervised task at training time ($\gamma = 0$) marginally increases the quality of detections, while subsequent finetuning iterations on the test sample substantially improve the number of correct detections while also decreasing both false positives and mislocalization errors.

{}

\vspace{-5mm}\subsubsection{Cross-task pseudo-labeling ablation} As explained in section \ref{sec:method} we have two options in the \our adaptation phase: either considering the whole image or focusing on pseudo-labeled bounding boxes
obtained from the detector after the first \our pretraining stage. For all our experiments we focused on the second case, indeed by
solving the auxiliary task only on objects, we limit the use of background features which may mislead the network towards solutions of the rotation task not based on relevant semantic information 
(\eg: finding fixed patterns in images, exploiting watermarks). We validate our choice by comparing it against using the rotation task on the entire image in both training and adaptation phases. Table $\ref{table:SS-Patch Ablation}$ shows results for VOC $\rightarrow$ AMD and Cityscapes $\rightarrow$ Foggy Cityscapes using \our. We observe that the choice of rotated regions is critical for the effectiveness of the algorithm. Solving the rotation task on objects using pseudo-annotations results in mAP improvements that range from $2.9$ to $5.9$ points, indicating that we learn better features for the main task.

\begin{table}[]
\centering\small
\begin{tabular}{c@{~~}c@{~~}c@{~~}}
%\toprule
\hline
%& Rotation on image & Rotation on objects \\ \hline
& $G_r(image)$ & $G_r(pseudoboxcrop)$ \\ \hline
VOC $\rightarrow$ Clipart  & 31.0 & \textbf{33.9} \\
VOC $\rightarrow$ Comic  & 21.0 & \textbf{26.9} \\
VOC $\rightarrow$ Watercolor  & 48.2 & \textbf{52.0} \\
Cityscapes $\rightarrow$ Foggy Cityscapes  & 27.7 & \textbf{31.9}\\
\hline
%\bottomrule
\end{tabular}
\caption{Rotating image vs rotating objects via pseudo-labeling on \our}
\label{table:SS-Patch Ablation}\vspace{-3mm}
\end{table}
\vspace{-5mm}\subsubsection{Self-supervised iterations} 
\begin{figure}[]
    \centering
    \includegraphics[width=0.9\textwidth]{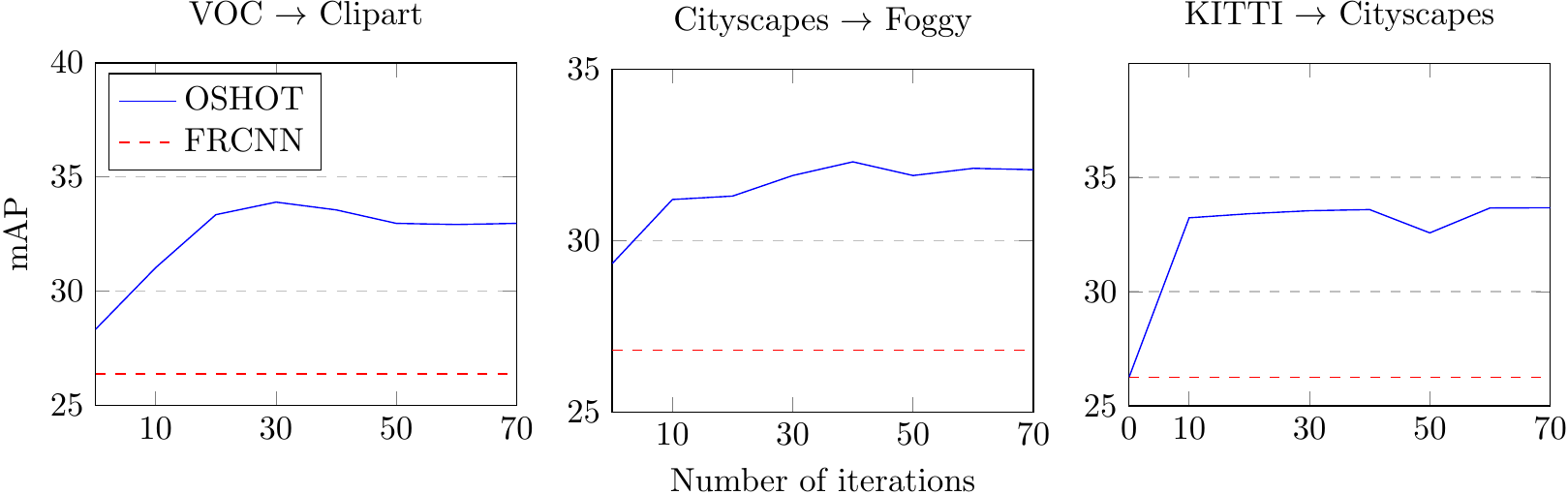}
    \caption{Performance of \our at different self-supervised iterations}
    \label{fig:Adaptive-Step}
\end{figure}

We study the effects of adaptating with up to $\gamma = 70$ iterations on VOC $\rightarrow$ Clipart, Cityscapes $\rightarrow$ FoggyCityscapes and KITTI $\rightarrow$ Cityscapes. Results are shown in figure \ref{fig:Adaptive-Step}. We observe a positive correlation between number of finetuning iterations and final mAP of the model in the earliest steps. This correlation is strong for the first 10 iterations, for which mAP increases spike on all observed targets. After about 30 iterations, performance tends to stabilize, indicating that increasing $\gamma$ beyond this point doesn't significantly alter final results.

{}

\vspace{-10mm}
\section{Conclusions}

This paper introduced for the first time \emph{one shot unsupervised cross-domain detection}, a scenario extremely relevant for the monitoring of image feeds on social media, where algorithms are called to adapt to a new visual domain from one single image. We showed that existing cross-domain detection methods suffer in this setting, as they are all explicitly designed to adapt from far larger quantities of target data. We presented the first deep architecture able to reduce the domain gap between source and target distribution by leveraging over one single target image. Our approach is based on a multi task structure that exploits self-supervision thanks to cross-task self-labeling. Extensive quantitative experiments and a qualitative analysis clearly demonstrate its effectiveness.

%\clearpage
%\clearpage
\bibliographystyle{splncs04}
\bibliography{egbib}

\section*{Appendix}
\appendix
\section{Full Ablation Results}
\subsubsection{Detection error analysis}
We complete here the detection error analysis that was only partially included in the main paper for space reasons. Specifically we consider all the three domain shift cases of VOC $\rightarrow$ AMD together with Cityscapes $\rightarrow$ Foggy Cityscapes, KITTI $\rightarrow$ Cityscapes and KITTI $\rightarrow$ Cityscapes. As reported in the main paper, for the first benchmark VOC $\rightarrow$ Clipart we follow \cite{hoiem2012diagnosing,diversify&match_Kim_2019_CVPR} considering the top 1k most confident detections and identifying three error types: correct (IoUgt $\geqslant$ 0.5), mislocalized (0.3 $\leqslant$ IoUgt $<$ 0.5) and background (IoUgt $<$ 0.3). For VOC $\rightarrow$ Comic and VOC $\rightarrow$ Watercolor we consider 2k most confident predictions, maintaining the same ratio of the first case given that the number of target samples are twice that of Clipart. A similar reasoning, that also takes care of the class cardinality, was applied to choose 6k most confident predictions for Foggy Cityscapes $\rightarrow$ Cityscapes, 1.5k for KITTI $\rightarrow$ Cityscapes and 20k for Cityscapes $\rightarrow$ KITTI.
\begin{figure}[tb]
    \centering
    \begin{tabular}{cc}
         \scriptsize{VOC $\rightarrow$ Clipart} & \hspace{2mm}\raisebox{-.5\height}{\includegraphics[width=0.75\textwidth]{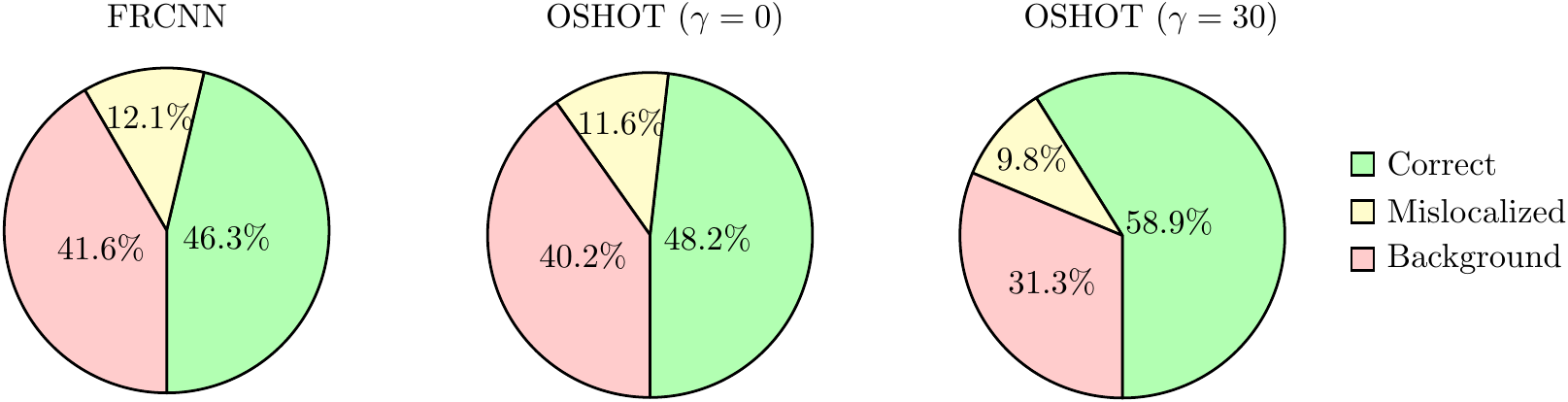}}\\
         \scriptsize{VOC $\rightarrow$ Comic} & \hspace{2mm}\raisebox{-.5\height}{\includegraphics[width=0.75\textwidth]{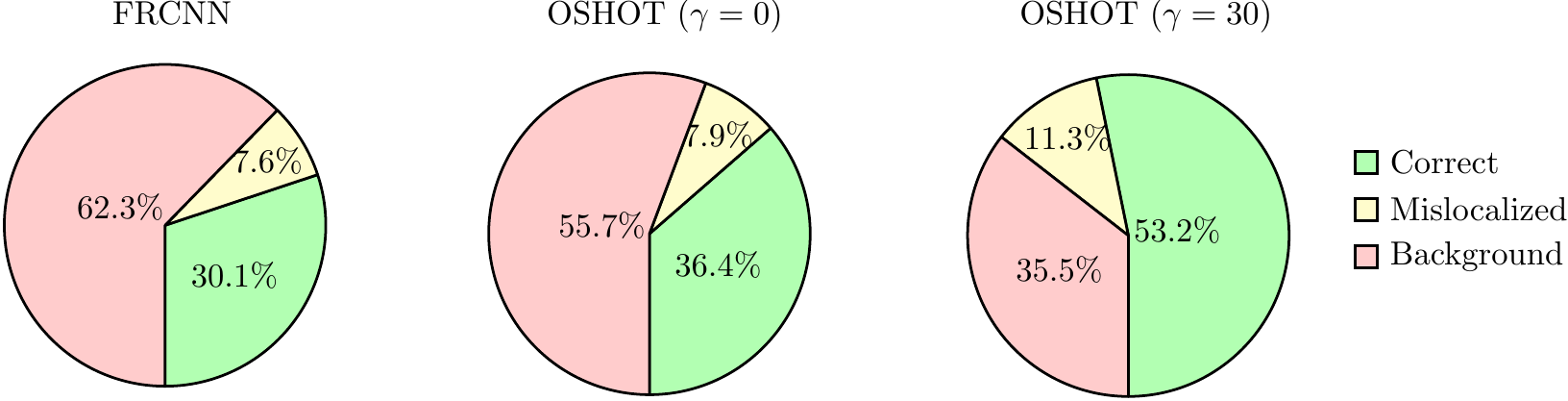}}\\
         \scriptsize{VOC $\rightarrow$ Watercolor} & \hspace{2mm}\raisebox{-.5\height}{\includegraphics[width=0.75\textwidth]{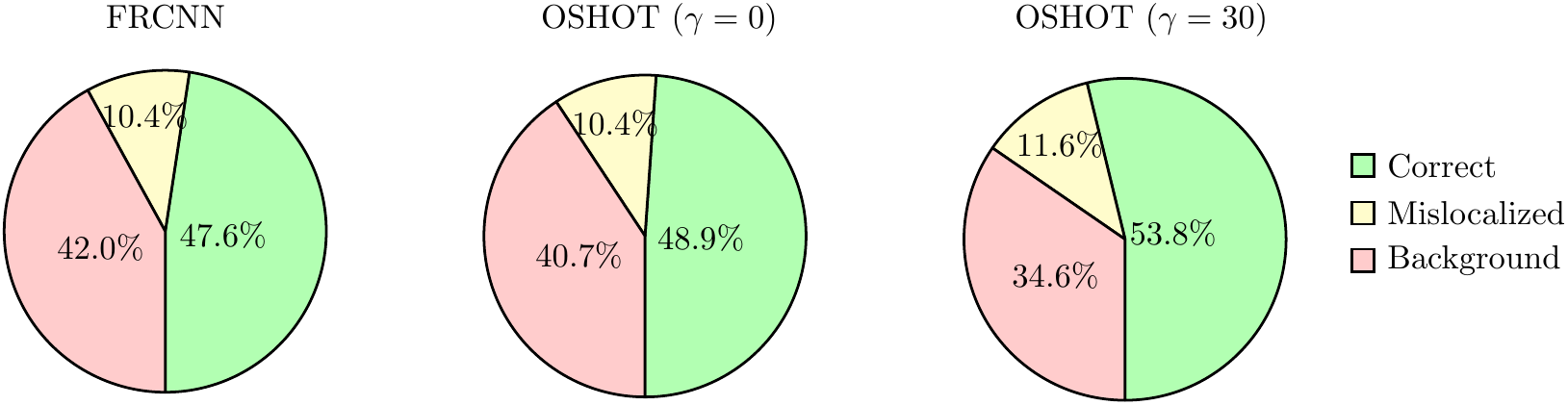}}\\
    \end{tabular}
    \caption{Detection error analysis on the three cases of VOC $\rightarrow$ AMD}
    \label{fig:comicdet}\vspace{-0.2cm}
\end{figure}
\begin{figure}[h!]
    \centering
    \begin{tabular}{cc}
         \scriptsize{Citys. $\rightarrow$ Foggy Citys.} &  \hspace{1mm}\raisebox{-.5\height}{\includegraphics[width=0.75\textwidth]{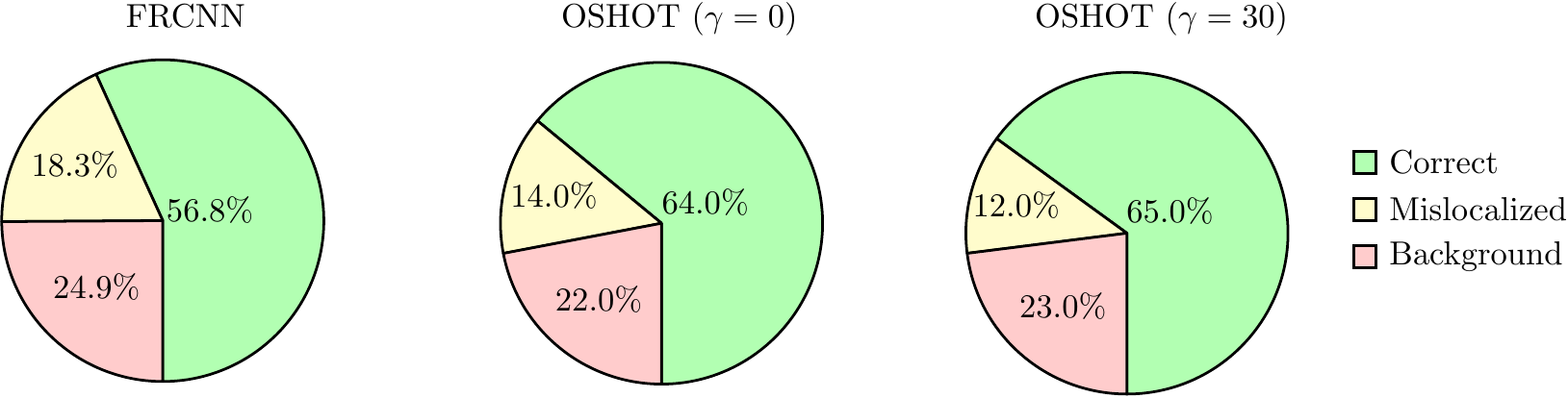}}\\ 
         \scriptsize{KITTI $\rightarrow$ Citys.} & \hspace{1mm}\raisebox{-.5\height}{\includegraphics[width=0.75\textwidth]{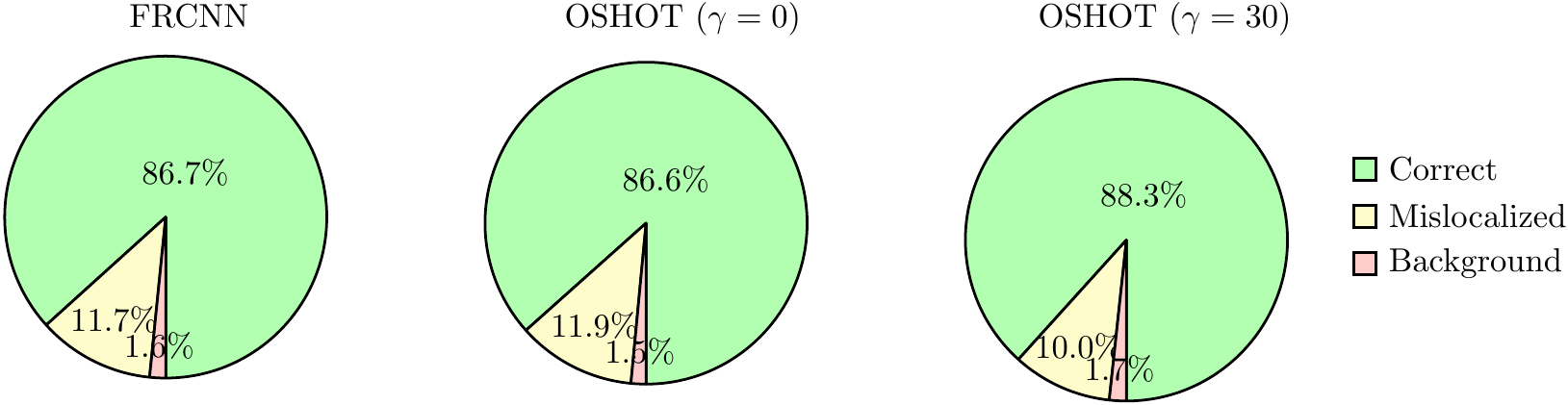}}\\
         \scriptsize{Citys. $\rightarrow$ KITTI} & \hspace{1mm}\raisebox{-.5\height}{\includegraphics[width=0.75\textwidth]{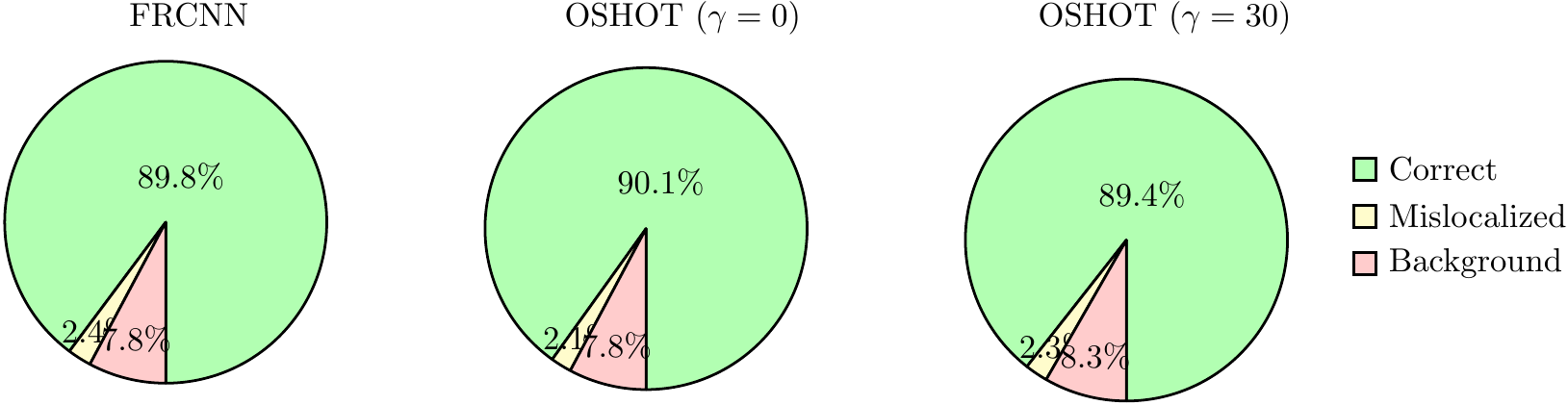}}\\
    \end{tabular}
    \caption{Detection error analysis on the three cases of urban scenes}
    \label{fig:urbandet}
\end{figure}
From Figure \ref{fig:comicdet} we can state that for both Clipart and Watercolor the advantage of adding the self-supervised task at training time is limited ($\gamma=0$), while the gain becomes evident when the number of adaptive iterations grows ($\gamma=30$). For Comic the improvement in performance appears already in the pretraining phase and further increases with adaptation. Overall the false positive errors decrease, while the ratio between the mislocalization error and correct localizations either decreases (Clipar, Comic) or remains stable (Watercolor).
A similar behaviour can be observed on the urban scenes, both when testing on Foggy Cityscapes and Cityscapes, as shown in the first two rows of Figure \ref{fig:urbandet}. For the last case of testing on KITTI, the results remain almost stable, confirming the same trend observed on the overall mAP performance discussed in the main paper. A neglegible drop of $0.7\%$ correct predictions appear when applying the adaptation phase for $\gamma=30$.

\begin{figure}[tb]
    \centering
    \includegraphics[width=0.8\textwidth]{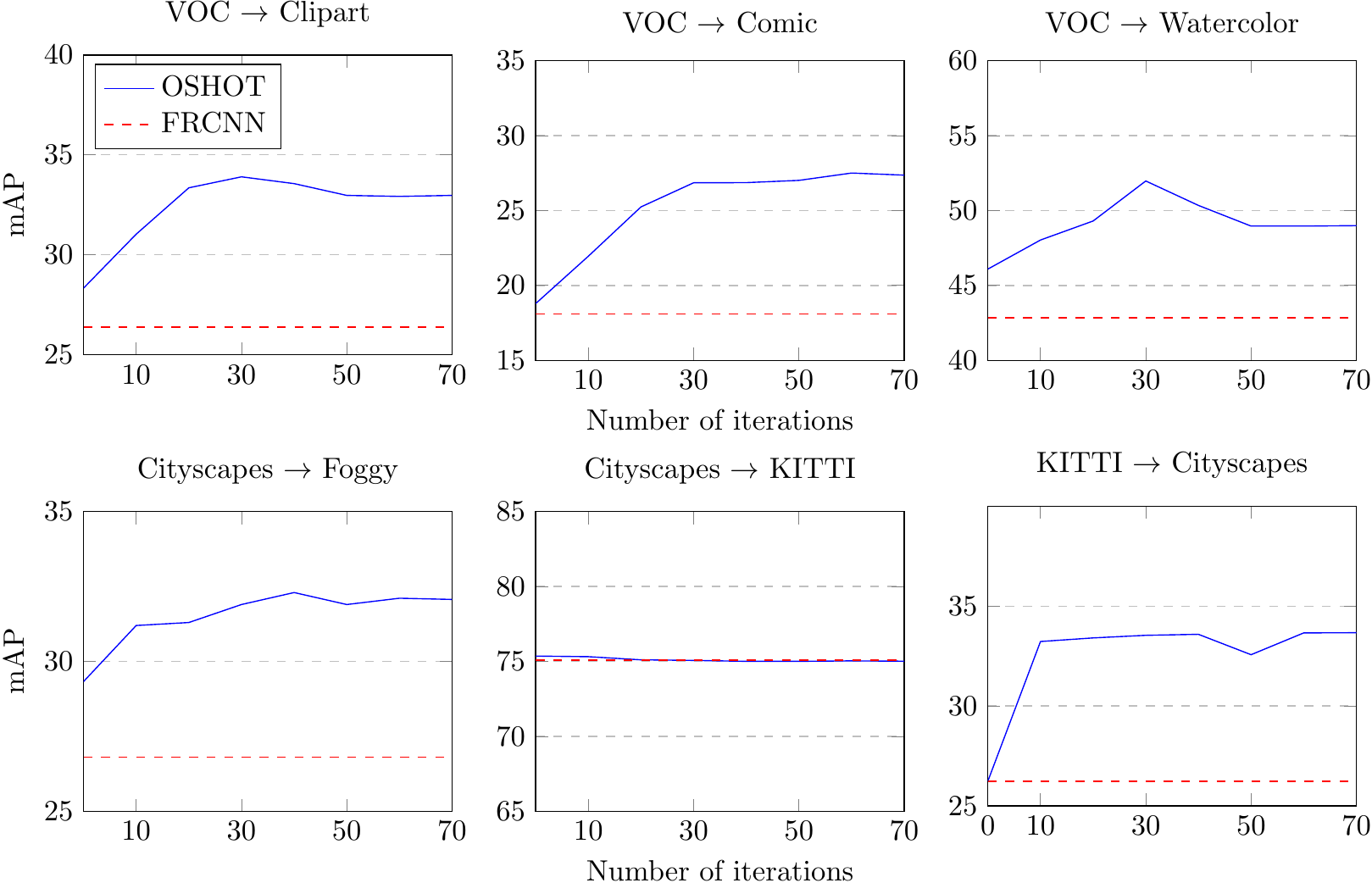}
    \caption{\our at different number of iterations for all testbeds}
    \label{fig:selfsupit}\vspace{-2mm}
\end{figure}{}
\subsubsection{Self-supervised iterations}
We report results of \our at different number of self-supervised iterations in Figure \ref{fig:selfsupit}.
We observe positive correlations between number of self-supervised iterations and final mAP on all targets except KITTI, for which final results are minimally affected by our adaptation procedure
(as well as by any other adaptive method used as reference - see Table 5 of the main paper).
The first 10 iterations show the most significant mAP change, while it gets to a stable plateau for further iterations.
\begin{figure}[h!]
    \centering
    \includegraphics[width=0.73\textwidth]{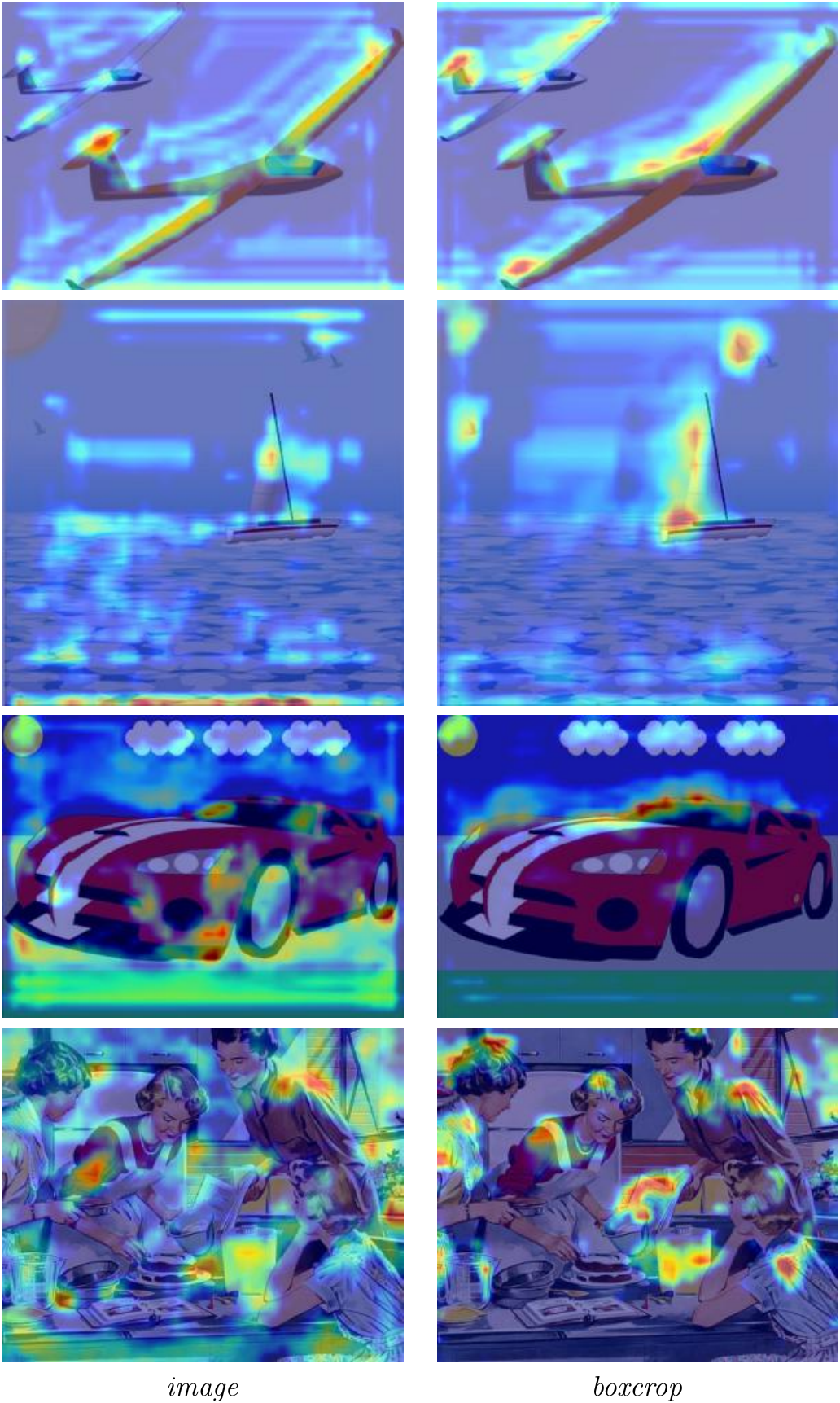}
    \caption{Visualization of the most relevant image regions produced by Grad-CAM when classifying the correct rotation with $G_r(image)$ and $G_r(boxcrop)$}
    \label{fig:qualitative1}\vspace{-2mm}
\end{figure}
\begin{figure}[tb]
    \centering
    \includegraphics[width=\textwidth]{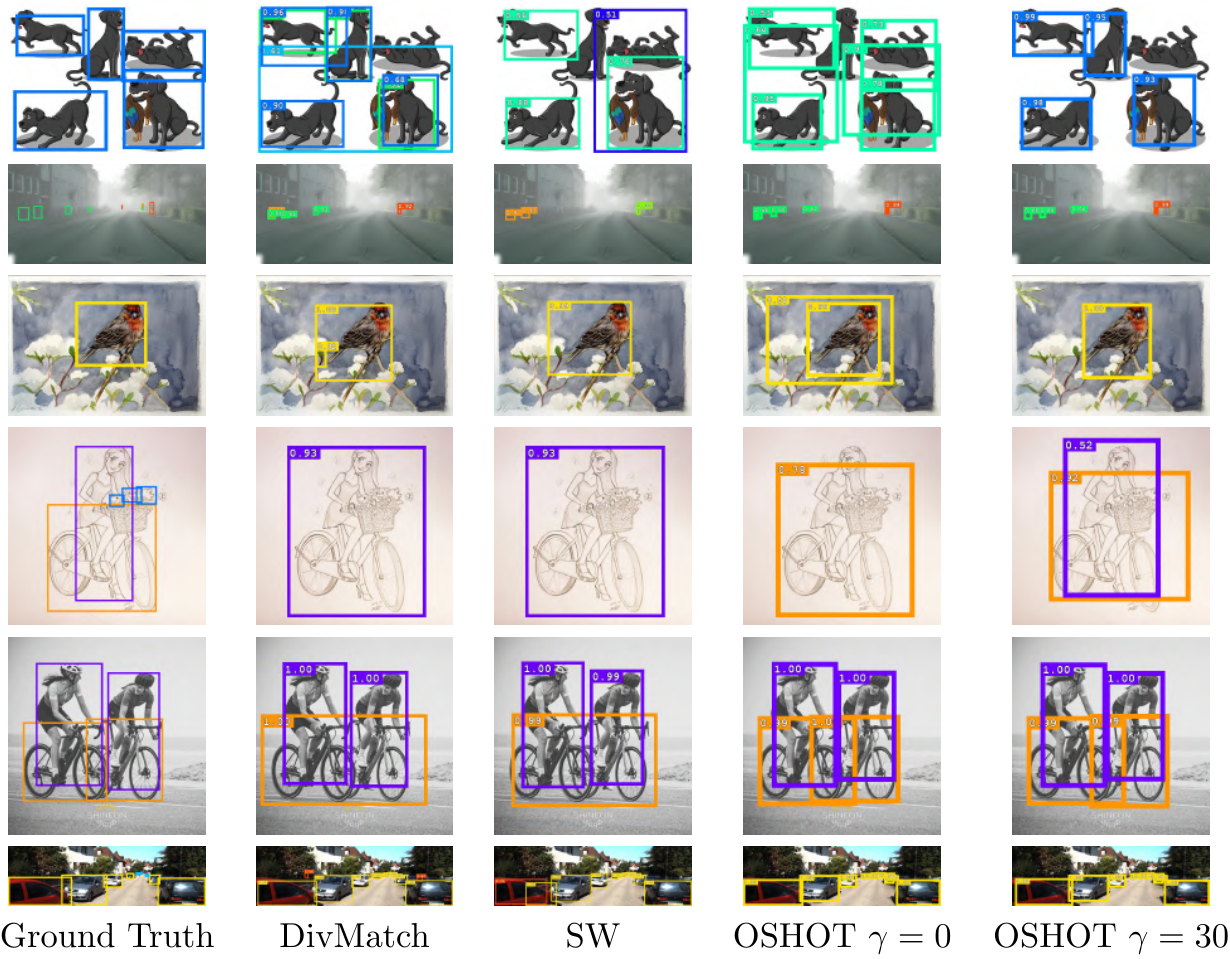}
    \caption{Qualitative visualization of detections with DivMatch, SW and \our on Comic (first row), Foggy Cityscapes (second row), Watercolor (third row), Social Bikes (fourth and fifth rows) and KITTI (sixth row). Numbers associated with bounding boxes indicate the detector's confidence}\vspace{-1mm}
    \label{fig:qualitative2}
\end{figure}
\section{Qualitative Analysis}
\subsubsection{Image vs Box rotation}
To validate our choice of considering box rotation over image rotation we set up a dedicated experiment. We ran the pretraining stage of \our on VOC by using either $G_r(image)$ or $G_r(boxcrop)$.
Then we tested the rotation classifier on whole images from the Clipart domain. In Figure \ref{fig:qualitative1} we show the results obtained with Grad-CAM \cite{gradcam_2017_ICCV} for the two cases, with heatmap indicating the most relevant regions responsible for recognizing the correct orientation. The Grad-CAM maps refer to the last output of the backbone feature extractor. We can see that, when the rotation classifier is trained on whole \textit{images} it learns to focus on the background (\eg the sky and the ground) in order to solve the task. On the contrary, when the \textit{boxcrop} operation is implemented to train the rotation classifier only on the relevant objects, it learns to look at objects' features even when it faces an entire image.

\subsubsection{Detection results of \our: baselines and self-supervised iterations}
Figure \ref{fig:qualitative2} shows some examples of detections of \our on images extracted from all the datasets considered in our work. We present as reference also the ground truth results as well as the predictions produced by DivMatch \cite{diversify&match_Kim_2019_CVPR} and SW \cite{Saito_2019_CVPR} that appear less precise than \our.
\begin{algorithm}[h!]
\label{alg:finetuning}
\SetAlgoLined
\KwData{$G_{f}$, $G_{d}$, $FC$, parameters $\theta_{f}$, $\theta_{r}$, $\theta_{d}$, rotator $R$, target image $x^{t}$}
$\theta_{f}^{*} \leftarrow \theta_{f}$ \\ 
$\theta_{r}^{*} \leftarrow \theta_{r}$ \\
 \While{still $\gamma$ iterations}{
  $b^{t},c^{t}$ $\leftarrow$ $G_{d}(G_{f}(x^{t}|\theta_{f}^{*})|\theta_{d})$\\
  $x_{r}^{t}$ $\leftarrow$ $R(x^{t})$\\
  $b_{r}^{t}$ $\leftarrow$ $R(b^{t})$\\
  minimize self-supervised loss $\mathcal{L}_{r}(FC_{\theta_{r}^{*}}(pseudoboxcrop(G_{f}(x_{r}^{t}|\theta_{f}^{*})|b_{r}^{t}))$\\
  update $\theta_{f}^{*}, \theta_{r}^{*}$ \\
 }
 predict label of test sample using $\mathcal{G}_{f}(\cdot|\theta_{f}^{*}), \mathcal{G}_{d}$
 \caption{\textbf{Adaptive phase of \our}}
 \label{alg:os}
\end{algorithm}
\section{\our pseudocode}
The pseudocode 
for the adaptive phase of \our is presented in Algorithm \ref{alg:os}. Here, $G_{f}$ and $G_{d}$ indicate the backbone feature extractor and detector, respectively parametrized by $\theta_{f}$ and $\theta_{d}$. $FC$ is the fully connected layer of the rotation classifier, parametrized by $\theta_{r}$, and $R$ is the rotator operator $R(x,\alpha)$ where $\alpha$, which indicates one random rotation to apply on $x$, is dropped for simplicity. The $pseudoboxcrop(\cdot|b)$ is an operator that applies cropping and ROI pooling on the input feature map based on the corresponding relative location of pseudo-boxes $b$.

\end{document}